\documentclass[runningheads]{llncs}
\usepackage[T1]{fontenc}
\usepackage{graphicx}
\usepackage[pagebackref,breaklinks,colorlinks]{hyperref}
\usepackage{graphicx}
\usepackage{amsmath}
\usepackage{multirow}
\usepackage{booktabs}
\usepackage{float}
\usepackage{amssymb}
\usepackage{svg}
\usepackage{physics}
\usepackage{marvosym}
\usepackage{subcaption}
\usepackage{orcidlink}

\usepackage{footnote}

\newcommand{\customfootnote}[1]{
    \begingroup
    \renewcommand{\thefootnote}{}
    \footnotetext{#1}
    \addtocounter{footnote}{-1}
    \endgroup
}

\begin{document}
\title{CAR: Contrast-Agnostic Deformable Medical Image Registration with Contrast-Invariant Latent Regularization}
\titlerunning{Contrast-Agnostic Deformable Medical Image Registration}
\author{Yinsong Wang\orcidlink{0009-0008-7288-4227} \and Siyi Du\orcidlink{0000-0002-9961-4533} \and Shaoming Zheng\orcidlink{0000-0001-5628-4311} \and Xinzhe Luo\orcidlink{0000-0003-2822-1633} \and Chen Qin\textsuperscript{(\Letter)}\orcidlink{0000-0003-3417-3092}}
%index{Yinsong, Wang}
%index{Siyi, Du}
%index{Shaoming, Zheng}
%index{Xinzhe, Luo}
%index{Chen, Qin}

%
\authorrunning{Y. Wang et al.}
\institute{Department of Electrical and Electronic Engineering \& I-X, Imperial College London, London, UK \\
\email{\{y.wang23,s.du23,s.zheng22,x.luo,c.qin15\}@imperial.ac.uk}}

\maketitle              % typeset the header of the contribution

\customfootnote{\textsuperscript{(\Letter)} Corresponding authors.}
\begin{abstract}

Multi-contrast image registration is a challenging task due to the complex intensity relationships between different imaging contrasts. Conventional image registration methods are typically based on iterative optimizations for each input image pair, which is time-consuming and sensitive to contrast variations. While learning-based approaches are much faster during the inference stage, due to generalizability issues, they typically can only be applied to the fixed contrasts observed during the training stage. In this work, we propose a novel contrast-agnostic deformable image registration framework that can be generalized to arbitrary contrast images, without observing them during training. Particularly, we propose a random convolution-based contrast augmentation scheme, which simulates arbitrary contrasts of images over a single image contrast while preserving their inherent structural information. To ensure that the network can learn contrast-invariant representations for facilitating contrast-agnostic registration, we further introduce contrast-invariant latent regularization (CLR) that regularizes representation in latent space through a contrast invariance loss. Experiments show that CAR outperforms the baseline approaches regarding registration accuracy and also possesses better generalization ability to unseen imaging contrasts. Code is available at \url{https://github.com/Yinsong0510/CAR}.

\end{abstract}
\section{Introduction}\label{sec1}
Multi-contrast images in magnetic resonance imaging (MRI) provide complementary information for characterizing tissues of human bodies, which are widely used in both qualitative and quantitative imaging in clinical diagnosis.
However, due to variations of acquisition protocols as well as possible patient movement during the acquisition process, the images are often spatially misaligned, which entails image registration for accurate downstream multi-contrast analysis and interpretation.
Registering images between different contrasts can be challenging due to the complex relationship between their intensity profiles.
Conventional approaches typically tackle this problem by optimizing information-theoretic similarity measures such as mutual information over the misaligned image pairs \cite{maes1997multimodality,maes2003medical}. 
Nevertheless, the iterative-based optimization framework is usually time-consuming and therefore can be limited in real-world applications.

Recently, deep learning-based registration approaches have shown their great potential for fast and accurate medical image registration. Several learning-based methods have also attempted to solve the multi-contrast image registration task. 
For instance, Qiu et al. \cite{qiu2021learning} have proposed to leverage the conventional mutual information as the similarity metric and embed it into an end-to-end learning-based registration framework.
% Mahapatra et al. \cite{mahapatra2018deformable} used generative adversarial networks (GAN) to deal with multi-contrast registration by using the generator network to output the deformation field and the warped moving image and the discriminator network to distinguish the warped moving image and the fixed image. 
An alternative way proposed by Qin et al. \cite{qin2019unsupervised} reduced the multi-contrast registration task to a mono-contrast one by disentangling images into shape and appearance latent spaces and aligning the image contrasts with style transfer approaches. Dey et al. \cite{dey2022contrareg} also proposed to use contrastive representation learning to learn a similarity loss between the warped and fixed multi-contrast images based on multi-scale PatchNCE using an autoencoder. 
Nevertheless, due to the network generalizability issues, these approaches can only deal with fixed imaging contrasts, which require the contrast of the image pairs to be the same during both training and inference stages.

To address the contrast variations between training and inference, there have also been some recent efforts in achieving contrast agnostic registration. Hoffmann et al. proposed to synthesize arbitrary contrast of images based on segmentation or randomly generated labels \cite{hoffmann2021synthmorph}. They used these generated images as the input to the registration network and optimized that based on original segmentation labels using soft dice loss \cite{dice1945measures}. However, this approach relies on the availability of segmentation maps and may not register images well in terms of fine-grained details, as the employed soft dice loss mainly focuses on the structure overlaps rather than pixel-level or volume-level differences.
Several recent works proposed training a neural network to learn a certain distance metric for multi-modal registration. 
Sideri et al. \cite{sideri2023mad} relied on modality and affine geometric augmentation for learning a contrast-agnostic distance measure based on image patch centers to tackle multi-contrast rigid registration.
Ronchetti et al. \cite{ronchetti2023disa} proposed to use a small CNN to learn a feature-level distance that approximates the conventional LC$^2$ similarity.
However, both approaches focused on similarity approximation, without explicitly enforcing contrast-agnostic learning.

% Several learning-based registration methods have been conducted to tackle the challenge of multi-contrast registration \cite{qiu2021learning,qin2019unsupervised,dey2022contrareg}. Unfortunately, all of these approaches can not generalize to unseen contrasts, which means they have to train a new model for every two contrasts of images they want to register. This has two limitations when we want to deploy our models in real-world scenarios. The first one is that training models for every two contrasts of the image is very time-consuming. This becomes even more serious when we need to register a MOLLI sequence that suffers from motion degradation to calculate the T1 mapping image, which typically contains 9 different contrasts, and therefore 8 models need to be trained. The second is that multi-contrast medical images are very scarce and difficult to obtain due to patient privacy. It is often unrealistic to obtain enough multi-contrast training data to train these models. Closest to our work, Hoffmann et al. \cite{hoffmann2021synthmorph} proposed to synthesize arbitrary contrast of MRI images based on segmentation maps and optimize the network based on soft dice loss. However, this approach relies on the availability of the segmentation maps and may not register images well in terms of fine-grained details because the soft dice loss merely focuses on the structure difference rather than pixel-level difference.

In this work, we propose a contrast-agnostic registration framework (CAR) for deformable image registration that only relies on single contrast images for training but can be generalized to other unseen contrasts.
Specifically, we propose a novel random convolution-based contrast-augmentation scheme to simulate arbitrary contrasts based on single contrast images. 
To encourage contrast-agnostic learning, we further enforce the network to learn contrast-invariant features by regularizing the feature representation in latent spaces using a proposed contrast invariance loss. 
Unlike existing learning-based multi-contrast registration frameworks \cite{dey2022contrareg,mahapatra2018deformable,qin2019unsupervised,qiu2021learning}, where each model can only register between fixed contrast images, CAR can train a single model that registers arbitrary contrasts of images without seeing the contrasts during training, which can significantly improve the generalizability and applicability for multi-contrast registration. In addition, CAR does not rely on segmentation maps for synthesis \cite{hoffmann2021synthmorph} or carefully designed networks for learning similarity losses \cite{ronchetti2023disa,sideri2023mad}. Instead, we can adopt simple mono-contrast similarity losses such as negative local normalized cross-correlation (LNCC) based on pre-augmented single contrast images, which enables the training of the network to be more lightweight and efficient. Experiments are performed on both brain MRI of T1 weighting of T2 weighting data and cardiac T1 mapping data with varied T1 weightings, and our results demonstrate the superior performance of CAR against baseline methods across various contrasts, even though the proposed model has not seen those contrasts during training.

\begin{figure}[!t]
	\centering
	\includegraphics[width=1.0\textwidth]{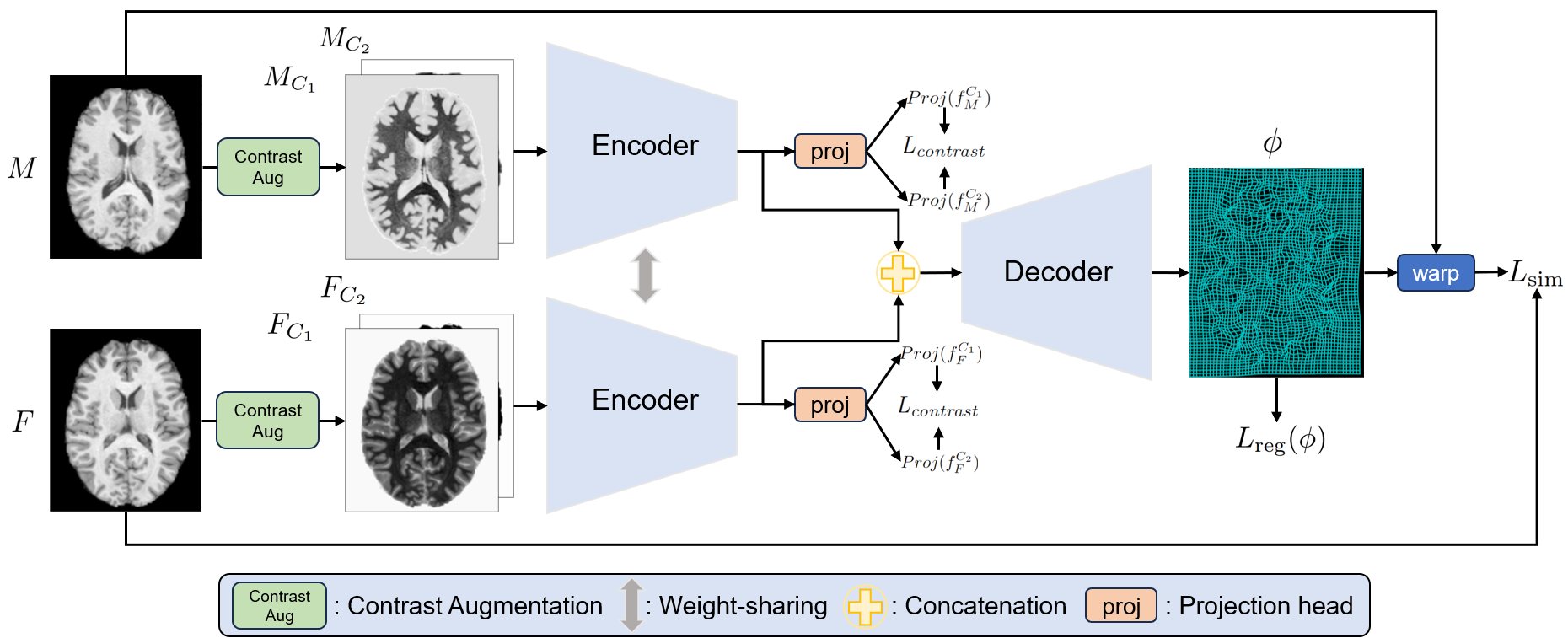} 
	\caption{Overview of the contrast-agnostic registration framework.
 The input fixed and moving image pair is fed to two Siamese encoders to produce contrast-invariant latent representations, which are taken by the decoder to predict the deformation field.
 } 
	\label{figure 2.1}
\end{figure}

\section{Methodology} 
Multi-contrast deformable image registration establishes dense and non-linear deformation field $\phi$ between a multi-contrast image pair, i.e. a moving image $M$ and a fixed image $F$. Learning-based approaches estimate the deformation field through a neural network $f_{\theta}$, $\phi=f_{\theta}(M, F)$, with loss functions typically consisting of a dissimilarity term $L_{\text{sim}}$ 
% which penalizes the difference between the warped moving image $M\circ\phi$ and the fixed image $F$
and a regularization term $L_{\text{reg}}$, 
% that penalizes the smoothness of the deformation field, 
i.e.,
\begin{equation}
\label{eq:1}
	L_{\text{total}} = L_{\text{sim}}(M\circ\phi, F) + \lambda L_{\text{reg}}(\phi),
\end{equation}
where $\lambda$ is a regularization hyperparameter. 
% Unfortunately, most learning-based multi-contrast registration approaches are limited to fixed contrasts that are available during training. By contrast, we propose a contrast-agnostic registration framework that can register any pair of contrasts even when the test contrasts are not observed during training. 
Conventional learning-based methods typically forecast the desired spatial transformation directly from image appearances, which not only renders the learned model contrast-dependent but disregards the fact that registration aims to find the intrinsic structural correspondence usually invariant to image contrasts.
To achieve contrast-agnostic registration that can deal with even unobserved contrasts as discussed in Section \ref{sec1}, in our method, we propose a random convolution-based contrast-augmentation scheme for contrast simulation (Section \ref{2.1}) and a contrast invariance loss for contrast-invariant representation learning (Section \ref{2.2}). Our overall framework is shown in Fig. \ref{figure 2.1}. 
% Finally, we introduce our registration network in section \ref{2.3}.
\subsection{Contrast Augmentation via Random Convolution}\label{2.1}
To enable a learning-based registration framework that can be contrast-agnostic, we first propose to simulate the variations of image contrasts by leveraging random convolution (RC) \cite{xu2021robust} for contrast augmentation. RC has been shown effective for learning robust representations for domain generalization in both the general computer vision field and medical imaging domain for tackling classification and segmentation tasks \cite{ouyang2022causality,sideri2023mad}. 
% However, it has been so far less explored in medical image registration problem, where its potential has yet to be uncovered in multi-contrast registration. 
Specifically, RC leverages randomly initialized convolution kernels to convolve with input images to adjust their appearances while preserving inherent shape information. In earlier works \cite{ouyang2022causality,xu2021robust}, it is shown that RC with large kernel size would introduce blurring effect, which can adversely impact the registration problem as it heavily relies on fine-grained details of images for learning pixel-level correspondences. To avoid this, we propose to set the kernel size to $1\times1$ to minimize the potential effects of the induced artifacts on structural details. This allows the contrast of similar tissues to vary homogeneously in the process, and this synchronizes with the rendering of multi-contrast images determined by imaging physics. Additionally, to ensure that the augmented images can maximally capture the variations of image contrasts, we further propose to stack multiple RC layers with LeakyReLu activation as proposed in \cite{xu2021robust} to model more complex and diverse contrast mappings, as well as simulating the non-linear intensity relationships as among different MRI contrasts. In each RC layer, the parameters of the convolution kernel are sampled independently from certain uniform distributions to enable the mapping to be diverse. The generated augmented images are then used as inputs to train our contrast-agnostic registration network. Illustrations of this contrast augmentation scheme and augmented images can be found in supplementary materials.

% Specifically, each convolutional layer comprises a $1\times1$ randomly initialized convolution kernel with a specific channel dimension for that layer. The randomly initialized kernel weights are sampled from a uniform distribution. Inspired by \cite{ouyang2022causality}, the pipeline is a shallow multi-layer convolutional network. The convolutional kernel of each layer is sampled from a uniform distribution $U(0, I)$ and then shifted to be zero-centered with a size of $1\times1$. Each layer has a different number of channels, the image is designed to increase its channel during intermediate layers and reduce its channel back at the final layer to produce the output. Since we use $1\times1$ convolution, this shallow network can also be regarded as a pixel-level MLP. After each layer's convolution, a LeakyReLu activation is added to provide non-linear intensity mapping to simulate non-linear intensity relationships between different imaging contrasts.

\subsection{Contrast-Invariant Latent Regularization} \label{2.2} 
To equip the network with the ability to learn contrast-invariant feature representations and stabilize the training process, we propose to regularize the feature representation in latent space using our proposed contrast invariance loss. Specifically, we assume that the latent representation of images that share the same shape information but with different contrasts should be close in the latent space. Therefore, we aim to pull the feature representations of the images with the same shape information to be close in latent space. Inspired by the contrastive learning-based methods \cite{chen2020simple,wang2021dense}, we also leverage a projection head to map the representation to a low dimensional space before computing the loss. Instead of using contrastive-based approaches, we propose to use a mean squared difference loss to calculate the difference of projected representations, which can provide a more direct constraint than the InfoNCE loss \cite{chen2020simple}. In detail, for each optimization step, we perform the network forward passes with two contrast-augmented moving-fixed image pairs $\{M_{C_{1}}, F_{C_{1}}\}$ and $\{M_{C_{2}}, F_{C_{2}}\}$. Through the encoder and the projection head, the latent representations of the two image pairs can then be obtained, i.e., $f_{M}^{C_{1}}, f_{F}^{C_{1}}, f_{M}^{C_{2}}, f_{F}^{C_{2}}$, with dimension $(C, N_x, N_y)$, where $C$ is the channel dimension and $(N_x, N_y)$ is the spatial dimension of the representation. Our contrast invariance loss function can be defined by

\begin{equation}
\scalebox{0.9}{
$\begin{aligned}
	{L_{\text{contrast}}} = \frac{1}{N_xN_y}\sum_{i=1}^{N_x}\sum_{j=1}^{N_y}&\ \bigg[({\operatorname{Proj}(f_{M}^{C_{1}}(i,j)) - \operatorname{Proj}(f_{M}^{C_{2}}(i,j))})^2 \\
  +&\ ({\operatorname{Proj}(f_{F}^{C_{1}}(i,j)) - \operatorname{Proj}(f_{F}^{C_{2}}(i,j))})^2\bigg],
\end{aligned}$
            }
\end{equation}
where $\operatorname{Proj}$ represents the projection head.  
% The reason why we regularize the feature representation in latent spaces is that latent spaces are in a low-dimensional manifold that captures the intrinsic feature of images, which could much be easier to optimize compared to the final deformation field that lies in a high-dimensional space with repeated information and complex patterns. 
Therefore, the contrast invariance loss essentially enforces the encoder to capture the intrinsic structural features that are invariant to various image contrasts or appearances \cite{achille2018emergence}.

\subsection{Overall Network Architecture and Loss Function} \label{2.3} The overall registration framework is illustrated in Fig. \ref{figure 2.1}. The registration network is a U-shape network \cite{balakrishnan2019voxelmorph} except we duplicate the encoders for moving images and fixed images respectively. Skip connections \cite{he2016deep} are adopted between the two Siamese encoders and the decoder. The moving image and the fixed image are taken as the input of their respective encoder and transformed to their corresponding latent representations $f_{M}$ and $f_{F}$, with $1/16$ of the original image size. The two latent representations are concatenated and fed into the decoder to output the final deformation field.

To train the registration network, our loss function is comprised of three terms: a dissimilarity loss $L_{\text{sim}}$, a regularization loss $L_{\text{reg}}$, and our proposed contrast invariance loss $L_{\text{contrast}}$. Since the input contrast-augmented image pair $\{M_{C_{i}}, F_{C_{i}}\}$ originated from the mono-contrast image pair $\{M, F\}$, we propose to use the pre-augmented mono-contrast image pairs for loss supervision with a mono-contrast dissimilarity measure. This therefore alleviates the need to fit any multi-contrast similarity loss through a neural network. In our work, we adopt the negative local normalized cross-correlation (LNCC) \cite{avants2008symmetric} as the dissimilarity, and a regularization loss formulated in the form of the L2-norm of the deformation field gradient. The overall loss function can be formulated as 

\begin{equation}
	\mathit{L_{\text{total}}} = -\operatorname{LNCC}(M\circ\phi, F) + \lambda_{1}||\nabla\phi||_{2}^{2} + \lambda_{2}\mathit{L_{\text{contrast}}},
\end{equation}
where $\lambda_{1}$ and $\lambda_{2}$ are the respective hyperparameters that control the strength of the regularization loss and contrast invariance loss. In this way, we will be able to achieve contrast-agnostic registration by jointly optimizing the deformation field and enforcing the network to learn contrast-invariant feature representations by our contrast-invariant latent regularization.

\begin{figure}[!t]
	\centering
\includegraphics[width=0.95\textwidth]{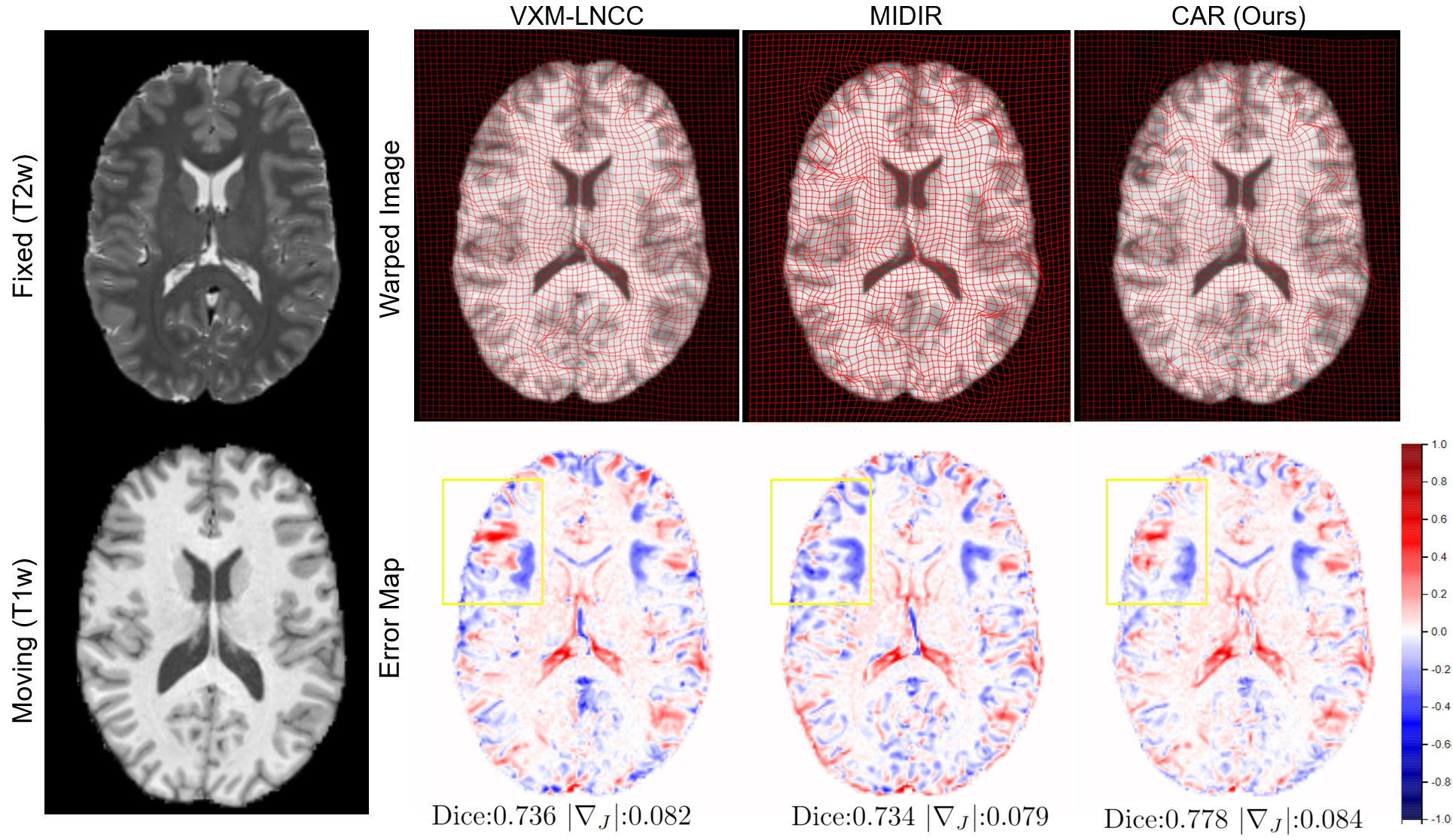} 
	\caption{Qualitative results on CamCAN dataset.} 
	\label{figure 3.1}
\end{figure}

\begin{figure}[!t]
	\centering
\includegraphics[width=0.95\textwidth]{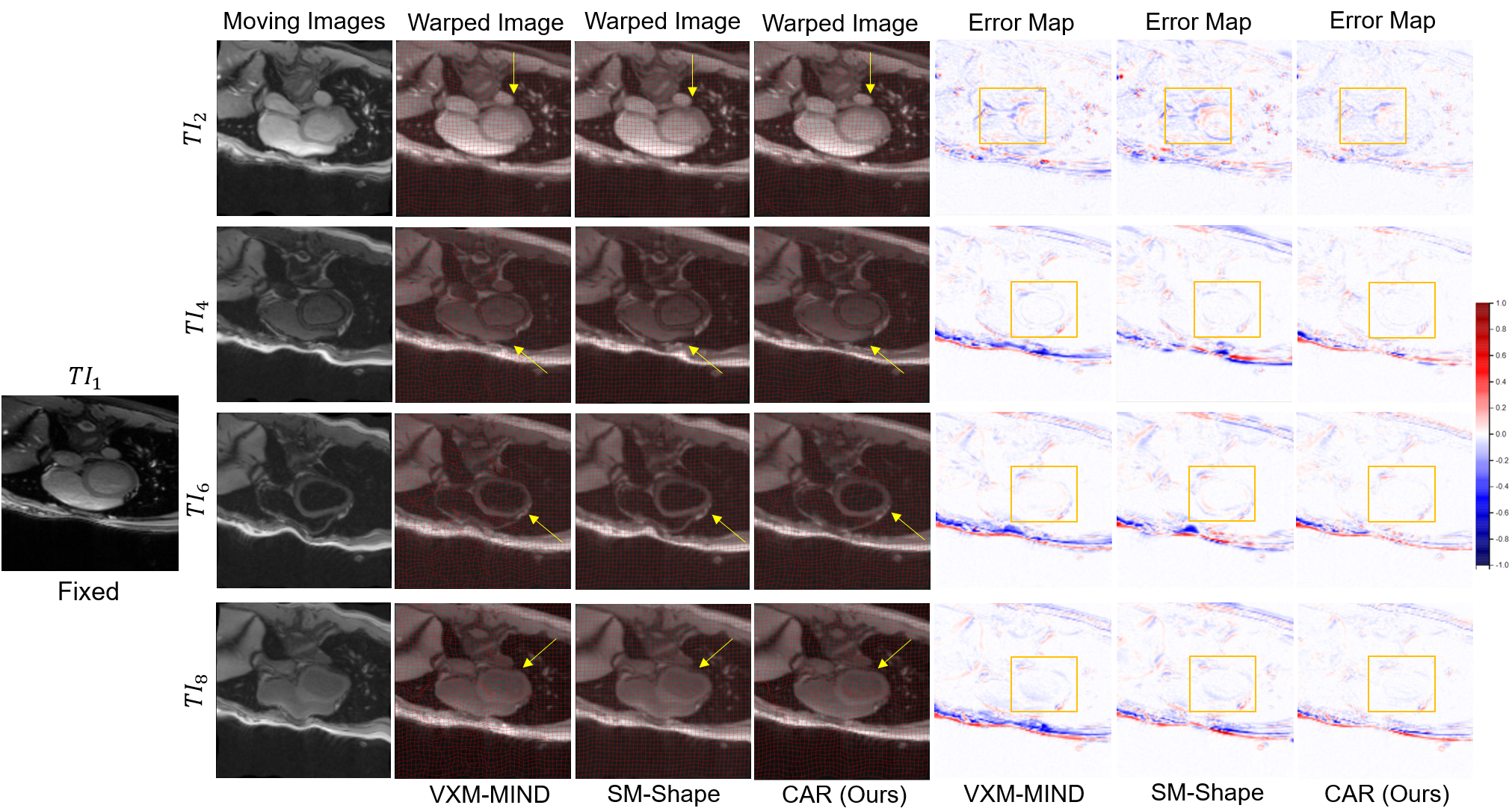} 
	\caption{Qualitative results on CMRxRecon dataset using T1 Mapping data for training.} 
	\label{figure 3.2}
\end{figure}

\section{Experiments and Results}
\noindent\textbf{Datasets.}
We evaluated CAR on two applications: (1) inter-subject 2D brain MRI registration of T1 weighted (T1w) - T2 weighted (T2w) images, and (2) intra-subject registration of multiple contrasts of T1 mapping acquisition in cardiac MRI. The 2D experiments on brain MRI registration mainly served as a proof-of-concept study to validate the method’s feasibility, and the cardiac datasets only contain 2D stacks of data for 2D experiments. For the brain registration task, we use T1w and T2w images of 652 subjects from the Cambridge Centre for Ageing and Neuroscience (CamCAN) project \cite{shafto2014cambridge,taylor2017cambridge}. We sampled the middle slice along the z-axis and cropped the sampled image into $160\times192$. The dataset was randomly split into 600, 10, and 42 volumes for training, validation, and testing, where subjects were randomly selected to form T1w-T2w image pairs. For the cardiac registration task, we used the public CMRxRecon cardiac datasets \cite{wang2021recommendation,wang2023cmrxrecon}. The CMRxRecon dataset consists of two parts: (1) Cine MRI, and (2) T1 mapping with images of nine different T1 weightings. Both Cine MRI and T1 Mapping data in the dataset contain 167 subjects. Each subject includes 7 to 12 or 4 to 5 short-axis view slices for Cine MRI data and T1 Mapping data respectively. 
% Each slice of Cine MRI data contains 12 images with the same T1 weighting to cover a cardiac cycle, whereas in T1 Mapping data each slice contains nine images with different T1 weightings. 
We cropped images in Cine and T1 Mapping data into $128\times128$. The dataset was randomly split into 137, 10, and 20 for training, validation, and testing. 
For the cardiac data, we simulated different random deformations to distort each image in the sequence using artificially generated FFDs with various mesh spacings following the approach from \cite{9965747}. For sequences within the same slice, we randomly selected an original image and a spatially distorted image as an image pair.

\begin{table}[!t]
\setlength{\tabcolsep}{1.5 mm}
\caption{Quantitative results of on both CamCAN and CMRxRecon datasets. * represents for values that CAR significantly outperformed with $p$-value $< 0.01$ in a paired $t$-test.}
\centering
\scalebox{0.58}{
\begin{tabular}{ccccccccccc}
  \toprule[1pt]
  \multirow{2}{*}{Methods} & \multicolumn{4}{c}{CamCAN} & \multicolumn{4}{c}{CMRxRecon} & \\
    \cmidrule(lr){2-9}
  
   	& Dice $\uparrow$ & $|\nabla_{J}|$ $\downarrow$ & $J_{<0}$\% $\downarrow$ & HD95 $\downarrow$ & Dice $\uparrow$ & $|\nabla_{J}|$ $\downarrow$ & $J_{<0}$\% $\downarrow$ & HD95 $\downarrow$ \\
	\cmidrule(lr){1-9}
 Unregistered & 0.519$\pm$0.109 & - & - & 5.418$\pm$1.829 & 0.815$\pm$0.066 & - & - &  2.445$\pm$0.562\\
 SyN & 0.707*$\pm$0.066 & \textbf{0.053}$\pm$0.009 & \textbf{0.461}$\pm$0.432\% & 3.368*$\pm$1.028 & 0.608*$\pm$0.295 & \textbf{0.046}$\pm$0.010 & 0.417*$\pm$0.478\% & 6.488*$\pm$5.238\\
    VXM-LNCC & 0.724*$\pm$0.056 & 0.085$\pm$0.009 & 0.472$\pm$0.163\% & 3.108*$\pm$0.685 & 0.859*$\pm$0.090 & 0.087*$\pm$0.018 & 0.108*$\pm$0.078\% & 2.114*$\pm$1.108\\
    VXM-MIND & 0.688*$\pm$0.061 & 0.108*$\pm$0.012 & 0.768*$\pm$0.288\% & 3.179*$\pm$0.703 & 0.863*$\pm$0.080 & 0.088*$\pm$0.012 & 0.166*$\pm$0.137\% & 1.888*$\pm$0.642\\
    MIDIR & 0.706*$\pm$0.045 & 0.068$\pm$0.006 & 0.480$\pm$0.216\% & 3.269*$\pm$0.599 & 0.763*$\pm$0.092 & 0.115*$\pm$0.015 & 0.491*$\pm$0.012\% & 3.286*$\pm$1.664\\
    SM-Shape & 0.715*$\pm$0.054 & 0.086$\pm$0.010 & 0.515*$\pm$0.208\% & 3.212*$\pm$0.785 & 0.868*$\pm$0.080 & 0.075*$\pm$0.007 & 0.110*$\pm$0.001\% & 1.854*$\pm$0.594\\
    \textbf{CAR (Ours)} & \textbf{0.743}$\pm$0.055 & 0.083$\pm$0.009 & 0.477$\pm$0.172\% & \textbf{3.040}$\pm$0.681 & \textbf{0.898}$\pm$0.068 & 0.047$\pm$0.005 & \textbf{0.002}$\pm$0.001\% & \textbf{1.514}$\pm$0.505\\
	\bottomrule[1pt]
\end{tabular}}
\label{table:quant}
\end{table}

% \begin{table}[!t]
% \setlength{\tabcolsep}{1.5 mm}
% \caption{Quantitative results on both CamCAM and CMRxRencon datasets. }
% \centering
% \scalebox{0.82}{
% \begin{tabular}{ccccccccccc}
%   \toprule[1pt]
% 	\multirow{2}{*}{Methods} & \multicolumn{3}{c}{CamCAN} & \multicolumn{3}{c}{CMRxRecon} & \multicolumn{3}{c}{CMRxRecon-Cine} & \\
 
%    \cmidrule(lr){2-10}
% 	& Dice & $|\nabla_{J}|$ & $J_{<0}$\% & Dice & $|\nabla_{J}|$ & $J_{<0}$\% & Dice & $|\nabla_{J}|$ & $J_{<0}$\% &\\
% 	\cmidrule(lr){1-11}
%  Unregistered & 0.519 & - & - & 0.815 & - & - & 0.815 & - & - &\\
%  SyN & 0.724 & 0.040 & 0.006\% & 0.815 & - & - & 0.815 & - & - &\\
%     VXM & 0.748 & 0.036 & 0.061\% & 0.851 & 0.061 & 0.013\% & 0.813 & \textbf{0.068} & \textbf{0.031\%}\\
%     VXM-MIND & 0.- & 0.- & - & 0.863 & 0.088 & 0.165\% & 0.730 & 0.132 & 1.68\%\\
%     MIDIR & 0.731 & \textbf{0.026} & 0.047\% & 0.763 & 0.115 & 0.491\% & 0.721 & 0.135 & 0.918\% \\
%     SM-Shape & 0.715 & 0.086 & 0.515\% & 0.868 & 0.075 & 0.110\% & 0.868 & 0.075 & 0.110\% &\\
%     \textbf{CAR (Ours)} & - & - & - & \textbf{0.898} & \textbf{0.047} & \textbf{0.002\%} & \textbf{0.893} & 0.080 & 0.043\%\\
% 	\bottomrule[1pt]
% \end{tabular}}
% \label{table:quant}
% \end{table}

%VXM (ch=256) & 0.724 & 0.085 & 0.472\% & 0.859 & 0.087 & 0.108\% & 0.808 & 0.080 & 0.950\% \\

\noindent\textbf{Evaluation metrics.}
We evaluated the registration accuracy and the deformation regularity of the registration. The registration accuracy was measured with Dice score between the segmentation of the warped moving image and the fixed image. To quantify the registration accuracy of the boundary of the structure of interest, we also measure the 95\% of the Hausdorff distance (HD95). We also evaluated the extent of extreme deformations, i.e., folding, by computing the ratio of points with negative Jacobian determinant ($J_{<0}\%$), and the smoothness of the deformation by computing the magnitude of the spatial gradient of Jacobian determinant $|\nabla_{J}|$ \cite{qiu2021learning}.

\noindent\textbf{Implementation details.}
The backbone network has two Siamese encoders and a decoder. The channel dimension of each convolutional layer in the two encoders is 128, and 256 for the decoder. The projection head for contrast-invariant latent regularization consists of a $1\times1$ convolution layer with the output channel set to 32. The random convolution-based contrast augmentation consists of 4 RC layers, where the kernel weights of each layer in section \ref{2.1} are sampled from a uniform distribution $U(0, 10)$. The kernel weights are reshifted to be zero-centered after sampling. A LeakyReLu activation with a negative slope of 0.2 is added after each RC layer. Experiments were conducted using Adam Optimizer on an NVIDIA A5000 GPU with a batch size of 16. The initial learning rate was set to be $1\times10^{-4}$ and started to decay to $1\times10^{-5}$ from the 5th epoch for convergence.

\begin{table}[!t]
\setlength{\tabcolsep}{1.5 mm}
\caption{Quantitative results of the generalizability experiment. The models are trained on CMRxRecon Cine data while tested on CMRxRecon T1 Mapping data. * represents for values that CAR significantly outperformed with $p$-value $< 0.01$ in a paired $t$-test.}
\centering
\scalebox{0.80}{
\begin{tabular}{ccccccc}
  \toprule[1pt]
   	Methods & Dice $\uparrow$ & $|\nabla_{J}|$ $\downarrow$ & $J_{<0}$\% $\downarrow$ & HD95 $\downarrow$\\
	\cmidrule(lr){1-5}
 Unregistered & 0.815$\pm$0.066 & - & - & 2.445$\pm$0.562\\
    VXM-LNCC & 0.808*$\pm$0.098 & 0.080$\pm$0.014 & 0.950*$\pm$0.287\% & 2.471*$\pm$0.890\\
    VXM-MIND & 0.730*$\pm$0.169 & 0.132*$\pm$0.030 & 1.678*$\pm$1.238 & 3.492*$\pm$1.903\\
    MIDIR & 0.721*$\pm$0.109 & 0.135*$\pm$0.027 & 0.918*$\pm$0.538\% & 3.489*$\pm$1.889\\
    SM-Shape & 0.868*$\pm$0.080 & \textbf{0.075}$\pm$0.007 & 0.110*$\pm$0.001\% & 1.854*$\pm$0.594 \\
    \textbf{CAR (Ours)} & \textbf{0.893}$\pm$0.067 & 0.080$\pm$0.008 & \textbf{0.043}$\pm$0.001\% & \textbf{1.567}$\pm$0.485\\
	\bottomrule[1pt]
\end{tabular}}
\label{table:quant_cine}
\end{table}

\noindent\textbf{Comparison study.}
CAR was first compared with a traditional iterative registration method SyN \cite{avants2008symmetric} with mutual information and a default Gaussian smoothing of 3 and three scales with 180, 80, 40 iterations, respectively following the setting of TransMorph \cite{chen2022transmorph}. We then compared CAR with SOTA learning-based methods, including VoxelMorph \cite{balakrishnan2019voxelmorph} using LNCC (VXM-LNCC) and MIND \cite{heinrich2012mind} (VXM-MIND) as the dissimilarity metrics respectively, a multi-contrast mutual information-based registration approach MIDIR \cite{qiu2021learning}, and a contrast-agnostic registration framework SynthMorph \cite{hoffmann2021synthmorph}. Note that the channel dimensions of all convolutional layers in VXM-LNCC, VXM-MIND, and SM-shape are set to 256 for a fair comparison with CAR. VoxelMorph and MIDIR were trained using all the available contrasts, i.e., both T1w and T2w images in brain MRI and all nine T1 weighting images in cardiac MRI. Synthetic images used for training SynthMorph were generated based on random label maps (SM-Shape). In contrast, CAR used only single contrast images for training, i.e., T1w images in brain MRI and $\mathrm{TI_{1}}$ images in cardiac MRI.

A quantitative comparison result on both brain MRI and cardiac MRI is summarized in Table \ref{table:quant}. For the cardiac registration task, the result is reported for registering misaligned images of $\mathrm{TI_{i=2:9}}$ to $\mathrm{TI_{1}}$. It can be observed that CAR achieves the best registration accuracy while possessing good deformation regularity for both tasks. For the brain MRI registration task, CAR demonstrates superior generalization ability as CAR only used T1w images for training but performed the best on the T1w-T2w registration task for all learning-based approaches in terms of registration accuracy and deformation regularity. This is similar to the cardiac registration task when our model was trained using only $\mathrm{TI_{1}}$ image but can be generalized to all other contrasts of images in T1 Mapping data. It can be noted that CAR has the lowest folding ratio. This is because we are dealing with nine different weightings of T1 images and the learned contrast-invariant representations may help the network get a more stable solution from images with different contrasts. Fig \ref{figure 3.1} and Fig \ref{figure 3.2} also show the qualitative comparison results of brain and cardiac registration tasks respectively, demonstrating that CAR can achieve low registration error and smooth deformation field. Both figures also show that CAR can register more fine-grained details than baseline approaches, which are shown in the highlighted regions.

To further evaluate the generalizability of CAR, we also trained our model and baseline methods on CMRxRecon Cine MRI data and evaluated model performance on the multi-contrast T1 Mapping data (also registering misaligned images of $\mathrm{TI_{i=2:9}}$ to $\mathrm{TI_{1}}$). The results are shown in Table \ref{table:quant_cine}. It can be shown all other methods except CAR and SM-shape failed to register the image with contrast they had not seen during training. In contrast, only CAR and SM-shape can still achieve good registration results. However, CAR has a significant increase in dice score and better deformation regularity than SM-shape, which demonstrates CAR's superior generalizability.

\begin{table}[!t]
\setlength{\tabcolsep}{1.5 mm}
\caption{Ablation Studies on the CamCAN dataset. * represents for values that CAR significantly outperformed with $p$-value $< 0.01$ in a paired $t$-test.}
\centering
\scalebox{0.80}{
\begin{tabular}{lllll}
  \toprule[1pt]
	 Methods & Dice $\uparrow$ & $|\nabla_{J}|$ $\downarrow$ & $J_{<0}\%$ $\downarrow$ & HD95 $\downarrow$ \\
   \cmidrule(lr){1-5}
   CAR w/o CLR & 0.733*$\pm$0.051 & 0.082$\pm$0.008 & 0.504*$\pm$0.158\% & 3.089*$\pm$0.641 \\
    CAR with RC-3 & 0.714*$\pm$0.054 & \textbf{0.082}$\pm$0.010 & 0.484$\pm$0.186\% & 3.129*$\pm$0.659\\
    CAR with InfoNCE &0.734*$\pm$0.057 & 0.084$\pm$0.010 & 0.501*$\pm$0.191\% & 3.053$\pm$0.642\\
    \textbf{CAR (Proposed)} & \textbf{0.743}$\pm$0.055 & 0.083$\pm$0.009 & \textbf{0.477}$\pm$0.172\% & \textbf{3.040}$\pm$0.681\\
    \bottomrule[1pt]
\end{tabular}}
\label{table:ablation}
\end{table}

\noindent\textbf{Ablation studies.}
To verify the effectiveness of CAR, we conducted ablation studies on the proposed random convolution-based contrast augmentation and CLR on the CamCAM datasets. We tested the effect of the random convolution kernel sizes on the model performance, where RC-3 represents that the kernel size is 3. We also replaced our proposed contrast invariance loss with InfoNCE loss \cite{chen2020simple} in contrastive learning to study its effectiveness. The results are summarized in Table \ref{table:ablation}. We can observe that any missing components of our model can lead to an inferior model performance from the results. Without CLR, the registration accuracy gets lower with a higher folding rate. Similar results can also be observed when replacing the proposed contrast invariance loss with InfoNCE loss in \cite{chen2020simple}. This means that CLR not only facilitates contrast-invariant representation learning but also improves the deformation regularity by producing a more plausible deformation field. Besides, a larger kernel size in the random convolution layer also leads to a performance drop. This is because a larger kernel size would lead to a blurring effect for the images, which can damage the voxel-to-voxel correspondences and therefore harm the registration task.

\section{Conclusion}

We present a novel contrast-agnostic deformable image registration framework (CAR) that can register arbitrary contrasts of images without observing them during training. To achieve this, we propose a random convolution-based approach to simulate variations of contrasts and develop a contrast invariance loss to regularize the feature representation in latent space for contrast-invariant feature representation learning. Experiments showed that on both tasks, CAR outperformed the state-of-the-art methods and demonstrated superior generalizability although CAR is contrast-agnostic. For future work, we will extend CAR to 3D applications as well as to more general multi-modal registration tasks.

\section*{Acknowledgement}
This work was partially supported by the Engineering and Physical Sciences Research Council [grant number EP/Y002016/1]. X. Luo was supported by the Engineering and Physical Sciences Research Council [grant number EP/X039277/1].

\bibliographystyle{splncs04}
\bibliography{mybibliography}

\begin{thebibliography}{10}
\providecommand{\url}[1]{\texttt{#1}}
\providecommand{\urlprefix}{URL }
\providecommand{\doi}[1]{https://doi.org/#1}

\bibitem{achille2018emergence}
Achille, A., Soatto, S.: Emergence of invariance and disentanglement in deep representations. The Journal of Machine Learning Research  \textbf{19}(1),  1947--1980 (2018)

\bibitem{avants2008symmetric}
Avants, B.B., Epstein, C.L., Grossman, M., Gee, J.C.: Symmetric diffeomorphic image registration with cross-correlation: evaluating automated labeling of elderly and neurodegenerative brain. Medical image analysis  \textbf{12}(1),  26--41 (2008)

\bibitem{balakrishnan2019voxelmorph}
Balakrishnan, G., Zhao, A., Sabuncu, M.R., Guttag, J., Dalca, A.V.: Voxelmorph: a learning framework for deformable medical image registration. IEEE transactions on medical imaging  \textbf{38}(8),  1788--1800 (2019)

\bibitem{basak2023semi}
Basak, H., Yin, Z.: Semi-supervised domain adaptive medical image segmentation through consistency regularized disentangled contrastive learning. In: International Conference on Medical Image Computing and Computer-Assisted Intervention. pp. 260--270. Springer (2023)

\bibitem{chen2022transmorph}
Chen, J., Frey, E.C., He, Y., Segars, W.P., Li, Y., Du, Y.: Transmorph: Transformer for unsupervised medical image registration. Medical image analysis  \textbf{82},  102615 (2022)

\bibitem{chen2020simple}
Chen, T., Kornblith, S., Norouzi, M., Hinton, G.: A simple framework for contrastive learning of visual representations. In: International conference on machine learning. pp. 1597--1607. PMLR (2020)

\bibitem{dey2022contrareg}
Dey, N., Schlemper, J., Salehi, S.S.M., Zhou, B., Gerig, G., Sofka, M.: Contrareg: Contrastive learning of multi-modality unsupervised deformable image registration. In: International Conference on Medical Image Computing and Computer-Assisted Intervention. pp. 66--77. Springer (2022)

\bibitem{dice1945measures}
Dice, L.R.: Measures of the amount of ecologic association between species. Ecology  \textbf{26}(3),  297--302 (1945)

\bibitem{he2016deep}
He, K., Zhang, X., Ren, S., Sun, J.: Deep residual learning for image recognition. In: Proceedings of the IEEE conference on computer vision and pattern recognition. pp. 770--778 (2016)

\bibitem{heinrich2012mind}
Heinrich, M.P., Jenkinson, M., Bhushan, M., Matin, T., Gleeson, F.V., Brady, M., Schnabel, J.A.: Mind: Modality independent neighbourhood descriptor for multi-modal deformable registration. Medical image analysis  \textbf{16}(7),  1423--1435 (2012)

\bibitem{hoffmann2021synthmorph}
Hoffmann, M., Billot, B., Greve, D.N., Iglesias, J.E., Fischl, B., Dalca, A.V.: Synthmorph: learning contrast-invariant registration without acquired images. IEEE transactions on medical imaging  \textbf{41}(3),  543--558 (2021)

\bibitem{9965747}
Luo, X., Zhuang, X.: $\mathcal {X}$-metric: An n-dimensional information-theoretic framework for groupwise registration and deep combined computing. IEEE Transactions on Pattern Analysis and Machine Intelligence  \textbf{45}(7),  9206--9224 (2023)

\bibitem{maes1997multimodality}
Maes, F., Collignon, A., Vandermeulen, D., Marchal, G., Suetens, P.: Multimodality image registration by maximization of mutual information. IEEE transactions on Medical Imaging  \textbf{16}(2),  187--198 (1997)

\bibitem{maes2003medical}
Maes, F., Vandermeulen, D., Suetens, P.: Medical image registration using mutual information. Proceedings of the IEEE  \textbf{91}(10),  1699--1722 (2003)

\bibitem{mahapatra2018deformable}
Mahapatra, D., Antony, B., Sedai, S., Garnavi, R.: Deformable medical image registration using generative adversarial networks. In: 2018 IEEE 15th International Symposium on Biomedical Imaging (ISBI 2018). pp. 1449--1453. IEEE (2018)

\bibitem{mcrobbie2017mri}
McRobbie, D.W., Moore, E.A., Graves, M.J., Prince, M.R.: MRI from Picture to Proton. Cambridge university press (2017)

\bibitem{messroghli2004modified}
Messroghli, D.R., Radjenovic, A., Kozerke, S., Higgins, D.M., Sivananthan, M.U., Ridgway, J.P.: Modified look-locker inversion recovery (molli) for high-resolution t1 mapping of the heart. Magnetic Resonance in Medicine: An Official Journal of the International Society for Magnetic Resonance in Medicine  \textbf{52}(1),  141--146 (2004)

\bibitem{ouyang2022causality}
Ouyang, C., Chen, C., Li, S., Li, Z., Qin, C., Bai, W., Rueckert, D.: Causality-inspired single-source domain generalization for medical image segmentation. IEEE Transactions on Medical Imaging  \textbf{42}(4),  1095--1106 (2022)

\bibitem{qin2019unsupervised}
Qin, C., Shi, B., Liao, R., Mansi, T., Rueckert, D., Kamen, A.: Unsupervised deformable registration for multi-modal images via disentangled representations. In: International Conference on Information Processing in Medical Imaging. pp. 249--261. Springer (2019)

\bibitem{qiu2022embedding}
Qiu, H., Hammernik, K., Qin, C., Chen, C., Rueckert, D.: Embedding gradient-based optimization in image registration networks. In: Medical Image Computing and Computer Assisted Intervention--MICCAI 2022: 25th International Conference, Singapore, September 18--22, 2022, Proceedings, Part VI. pp. 56--65. Springer (2022)

\bibitem{qiu2021learning}
Qiu, H., Qin, C., Schuh, A., Hammernik, K., Rueckert, D.: Learning diffeomorphic and modality-invariant registration using b-splines. In: Medical Imaging with Deep Learning (2021)

\bibitem{ronchetti2023disa}
Ronchetti, M., Wein, W., Navab, N., Zettinig, O., Prevost, R.: Disa: Differentiable similarity approximation for universal multimodal registration. In: International Conference on Medical Image Computing and Computer-Assisted Intervention. pp. 761--770. Springer (2023)

\bibitem{rueckert1999nonrigid}
Rueckert, D., Sonoda, L.I., Hayes, C., Hill, D.L., Leach, M.O., Hawkes, D.J.: Nonrigid registration using free-form deformations: application to breast mr images. IEEE transactions on medical imaging  \textbf{18}(8),  712--721 (1999)

\bibitem{shafto2014cambridge}
Shafto, M.A., Tyler, L.K., Dixon, M., Taylor, J.R., Rowe, J.B., Cusack, R., Calder, A.J., Marslen-Wilson, W.D., Duncan, J., Dalgleish, T., et~al.: The cambridge centre for ageing and neuroscience (cam-can) study protocol: a cross-sectional, lifespan, multidisciplinary examination of healthy cognitive ageing. BMC neurology  \textbf{14},  1--25 (2014)

\bibitem{shi2022xmorpher}
Shi, J., He, Y., Kong, Y., Coatrieux, J.L., Shu, H., Yang, G., Li, S.: Xmorpher: Full transformer for deformable medical image registration via cross attention. In: International Conference on Medical Image Computing and Computer-Assisted Intervention. pp. 217--226. Springer (2022)

\bibitem{sideri2023mad}
Sideri-Lampretsa, V., Zimmer, V.A., Qiu, H., Kaissis, G., Rueckert, D.: Mad: Modality agnostic distance measure for image registration. In: International Conference on Medical Image Computing and Computer-Assisted Intervention. pp. 147--156. Springer (2023)

\bibitem{taylor2017cambridge}
Taylor, J.R., Williams, N., Cusack, R., Auer, T., Shafto, M.A., Dixon, M., Tyler, L.K., Henson, R.N., et~al.: The cambridge centre for ageing and neuroscience (cam-can) data repository: Structural and functional mri, meg, and cognitive data from a cross-sectional adult lifespan sample. neuroimage  \textbf{144},  262--269 (2017)

\bibitem{thirion1998image}
Thirion, J.P.: Image matching as a diffusion process: an analogy with {Maxwell}'s demons. Medical image analysis  \textbf{2}(3),  243--260 (1998)

\bibitem{wang2021recommendation}
Wang, C., Li, Y., Lv, J., Jin, J., Hu, X., Kuang, X., Chen, W., Wang, H.: Recommendation for cardiac magnetic resonance imaging-based phenotypic study: imaging part. Phenomics  \textbf{1},  151--170 (2021)

\bibitem{wang2023cmrxrecon}
Wang, C., Lyu, J., Wang, S., Qin, C., Guo, K., Zhang, X., Yu, X., Li, Y., Wang, F., Jin, J., et~al.: Cmrxrecon: an open cardiac mri dataset for the competition of accelerated image reconstruction. arXiv preprint arXiv:2309.10836  (2023)

\bibitem{wang2021dense}
Wang, X., Zhang, R., Shen, C., Kong, T., Li, L.: Dense contrastive learning for self-supervised visual pre-training. In: Proceedings of the IEEE/CVF Conference on Computer Vision and Pattern Recognition. pp. 3024--3033 (2021)

\bibitem{10230464}
Wang, Y., Qiu, H., Qin, C.: Conditional deformable image registration with spatially-variant and adaptive regularization. In: 2023 IEEE 20th International Symposium on Biomedical Imaging (ISBI). pp.~1--5 (2023)

\bibitem{xu2021robust}
Xu, Z., Liu, D., Yang, J., Raffel, C., Niethammer, M.: Robust and generalizable visual representation learning via random convolutions. In: International Conference on Learning Representations (2021)

\end{thebibliography}




% LNCS Homepage, \url{http://www.springer.com/lncs}, last accessed 2023/10/25
% \end{thebibliography}
\end{document}